\documentclass[acmtog]{acmart}

\setcopyright{acmlicensed}
\acmJournal{TOG}
\acmYear{2024} 
\acmVolume{43} 
\acmNumber{4} 
\acmArticle{121} 
\acmMonth{7}
\acmDOI{10.1145/3658205}

\usepackage{booktabs} 
\usepackage{makecell}
\usepackage{subfigure}
\usepackage{graphicx}
\usepackage[misc]{ifsym}
\usepackage{xcolor}         
\usepackage{gensymb}

\definecolor{ao}{rgb}{0.0, 0.0, 1.0}
\definecolor{airforceblue}{rgb}{0.36, 0.54, 0.66}
\definecolor{ceruleanblue}{rgb}{0.16, 0.32, 0.75}
\definecolor{cerulean}{rgb}{0.0, 0.48, 0.65}
\definecolor{celestialblue}{rgb}{0.29, 0.59, 0.82}
\definecolor{azure(colorwheel)}{rgb}{0.0, 0.5, 1.0}
\definecolor{babyblue}{rgb}{0.54, 0.81, 0.94}
\definecolor{babyblueeyes}{rgb}{0.63, 0.79, 0.95}
\definecolor{ballblue}{rgb}{0.13, 0.67, 0.8}

\definecolor{asparagus}{rgb}{0.53, 0.66, 0.42}
\definecolor{ao(english)}{rgb}{0.0, 0.5, 0.0}
\definecolor{applegreen}{rgb}{0.55, 0.71, 0.0}
\definecolor{armygreen}{rgb}{0.29, 0.33, 0.13}
\definecolor{gray-asparagus}{rgb}{0.27, 0.35, 0.27}
\definecolor{green(ryb)}{rgb}{0.4, 0.69, 0.2}

\definecolor{amethyst}{rgb}{0.6, 0.4, 0.8}
\definecolor{antiquefuchsia}{rgb}{0.57, 0.36, 0.51}
\definecolor{blue-violet}{rgb}{0.54, 0.17, 0.89}
\definecolor{brightlavender}{rgb}{0.75, 0.58, 0.89}
\definecolor{brightube}{rgb}{0.82, 0.62, 0.91}
\definecolor{brilliantlavender}{rgb}{0.96, 0.73, 1.0}

\definecolor{amber}{rgb}{1.0, 0.75, 0.0}
\definecolor{amber(sae/ece)}{rgb}{1.0, 0.49, 0.0}
\definecolor{atomictangerine}{rgb}{1.0, 0.6, 0.4}
\definecolor{burntorange}{rgb}{0.8, 0.33, 0.0}
\definecolor{burntsienna}{rgb}{0.91, 0.45, 0.32}
\definecolor{cadmiumorange}{rgb}{0.93, 0.53, 0.18}
\definecolor{carrotorange}{rgb}{0.93, 0.57, 0.13}
\definecolor{chocolate(web)}{rgb}{0.82, 0.41, 0.12}
\definecolor{chromeyellow}{rgb}{1.0, 0.65, 0.0}
\definecolor{darkorange}{rgb}{1.0, 0.55, 0.0}
\definecolor{darktangerine}{rgb}{1.0, 0.66, 0.07}
\definecolor{deepcarrotorange}{rgb}{0.91, 0.41, 0.17}
\definecolor{deepsaffron}{rgb}{1.0, 0.6, 0.2}
\definecolor{fulvous}{rgb}{0.86, 0.52, 0.0}

\def\sysname{TIP-Editor}
\def\step1{stepwise 2D personalization strategy}

\def\dreambooth3d{D-DreamFusion*}
\def\nerf2nerf{Instruct-N2N}

\usepackage[ruled]{algorithm2e} % For algorithms

\SetAlFnt{\small}
\SetAlCapFnt{\small}
\SetAlCapNameFnt{\small}
\SetAlCapHSkip{0pt}

\usepackage{enumitem}
\setlist{topsep=0pt, leftmargin=*}

\begin{document}

\title{TIP-Editor: An Accurate 3D Editor Following Both Text-Prompts And Image-Prompts}

\author{Jingyu Zhuang}
\authornote{Work done during an internship at Tencent AI Lab.}
\email{zhuangjy6@mail2.sysu.edu.cn}
\affiliation{%
  \institution{Sun Yat-sen University}
  \country{China and }
  \institution{Tencent AI Lab}
  \country{China}
}

\author{Di Kang}
\email{di.kang@outlook.com}
\affiliation{%
  \institution{Tencent AI Lab}
  \country{China}
}

\author{Yan-Pei Cao}
\email{caoyanpei@gmail.com}
\affiliation{%
  \institution{Tencent AI Lab}
  \country{China}
}

\author{Guanbin Li}
% \authornote{Corresponding author. Welcome to \href{https://www.sysu-hcp.net/projects/cv/127.html}{\textcolor{red}{\emph{Project page}}} }
\authornote{Corresponding author. Welcome to \href{https://zjy526223908.github.io/TIP-Editor/}{\textcolor{red}{\emph{Project page}}} }
\email{liguanbin@mail.sysu.edu.cn}
\affiliation{%
  \institution{Sun Yat-sen University}
  \country{China and }
  \institution{Peng Cheng Laboratory}
  \country{China}
}

\author{Liang Lin}
\email{linliang@ieee.org}
\affiliation{%
  \institution{Sun Yat-sen University}
  \country{China and }
  \institution{Peng Cheng Laboratory}
  \country{China}
}

\author{Ying Shan}
\email{yingsshan@tencent.com}
\affiliation{%
  \institution{Tencent AI Lab}
  \country{China}
}

\begin{abstract}
Text-driven 3D scene editing has gained significant attention owing to its convenience and user-friendliness.
However, existing methods still lack accurate control of the specified appearance and location of the editing result due to the inherent limitations of the text description.
To this end, we propose a 3D scene editing framework, \sysname{},
that accepts both \textbf{\underline{t}}ext and \textbf{\underline{i}}mage \textbf{\underline{p}}rompts and a 3D bounding box to specify the editing region.
With the image prompt, users can conveniently specify the detailed appearance/style of the target
content in complement to the text description, enabling accurate control of the appearance.
Specifically, \sysname{} employs a stepwise 2D personalization strategy to better learn the representation of the existing scene and the reference image,
in which a localization loss is proposed to encourage correct object placement as specified by the bounding box.
Additionally, \sysname{} utilizes explicit and flexible 3D Gaussian splatting (GS) as the 3D representation to facilitate local editing while keeping the background unchanged.
Extensive experiments have demonstrated that
\sysname{} conducts accurate editing following the text and image prompts in the specified bounding box region, consistently outperforming the baselines in editing quality, and the alignment to the prompts, qualitatively and quantitatively.
\end{abstract}

%
% The code below should be generated by the tool at
% http://dl.acm.org/ccs.cfm
% Please copy and paste the code instead of the example below.
%


\begin{teaserfigure}
\centering
\includegraphics[width=1.0\textwidth]{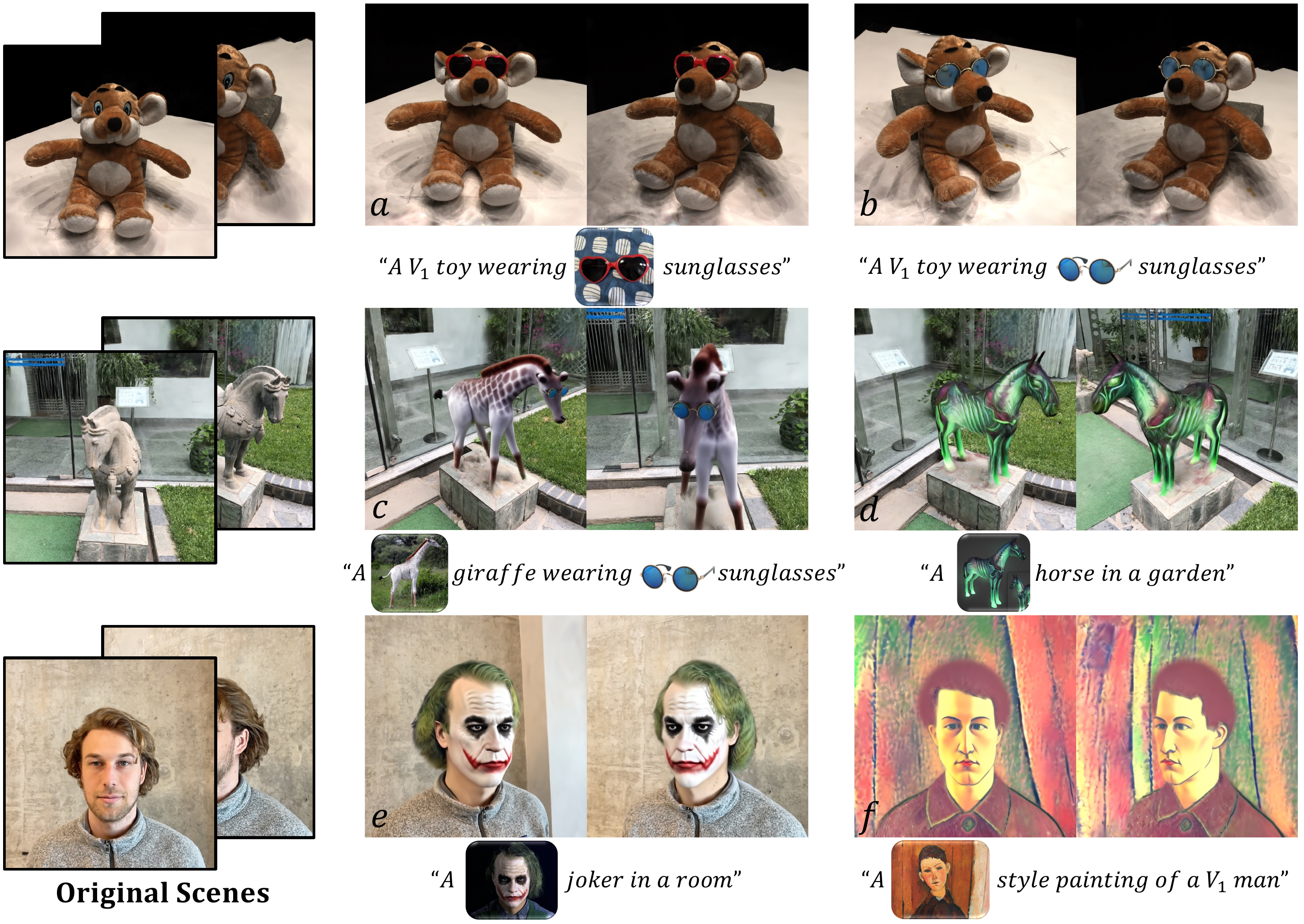}
\caption{
\sysname{} excels in precise and high-quality localized editing given a 3D bounding box,
and allows the users to perform various types of editing on a 3D scene, 
such as object insertion $(a, b)$, whole object replacement $(d)$, part-level object editing $(e)$,
combination of these editing types (i.e. sequential editing, $(c)$), and stylization $(f)$.
The editing process is guided by not only the text but also one reference image, 
which serves as the complement of the \textit{textual} description and results in more accurate editing control.
Images in the text prompts denote their associated \emph{rare tokens}, which are fixed without optimization.
}
\label{fig:teaser}
\end{teaserfigure}

\maketitle

\section{Introduction}

Due to the unprecedented photorealistic rendering quality, methods that use radiance field-related representations (e.g. NeRF~\cite{mildenhall2021nerf} and 3D Gaussian Splatting~\cite{kerbl20233d}) have been more and more popular in 3D reconstruction field~\cite{yu2021pixelnerf,li2021mine,cao2023scenerf} and various downstream 3D editing tasks, such as texture editing~\cite{xiang2021neutex, richardson2023texture},
shape deformation~\cite{yang2022neumesh, xu2022deforming},
scene decomposition~\cite{tang2022compressible},
and stylization~\cite{wang2023nerf}.

Generative editing, which only requires high-level instructions (e.g. text prompts), emerges as a new approach in complement to previous painting-like and sculpting-like editing approaches~\cite{xiang2021neutex,yang2022neumesh} that require \emph{extensive} user interactions.
Among these methods, text-driven methods~\cite{haque2023instruct,zhuang2023dreameditor} 
have gained significant attention due to their convenience and have achieved remarkable progress due to the success of large-scale text-to-image (T2I) models.

However, 
methods using only text as the condition struggle to precisely generate editing results with the specified appearance at the specified location due to the inherent limitations of the text description.
For example, existing text-driven methods usually produce less satisfactory results (Fig.~\ref{fig_visual_compare}) if we want to dress the toy in a special heart-shaped sunglasses or give the male the Joker makeup appeared in the movie \emph{The Dark Knight}.
Moreover, it is hard to specify the accurate editing location by text guidance (Fig.~\ref{fig_ablation_learning_1}).
These challenges primarily stem from the diverse appearances of the generated objects and the diverse spatial layout of the generated scenes.

To overcome the challenges above,
we present \sysname{}, which allows the users to intuitively, conveniently, and accurately edit the exiting GS-based (3D Gaussian Splatting~\cite{kerbl20233d}) radiance fields using both \textbf{\underline{t}}ext prompts and \textbf{\underline{i}}mage \textbf{\underline{p}}rompts.
Our framework achieves such capabilities through two crucial designs.
(1) The first one is a novel stepwise 2D personalization strategy that enables precise appearance control (via a reference image) and location control (via a 3D bounding box).
Specifically, it contains a scene personalization step, which includes a localization loss to ensure the editing occurs inside the user-defined editing region, and a separate novel content personalization step dedicated to the reference image based on LoRA~\cite{hu2021lora}.
(2) The second one is adopting explicit and flexible 3D Gaussian splatting~\cite{kerbl20233d} as the 3D representation since it is efficient and, more importantly, highly suitable for local editing.

We conduct comprehensive evaluations of \sysname{} across various real-world scenes, including objects, human faces, and outdoor scenes. 
Our editing results (Fig.~\ref{fig:teaser} and Fig.~\ref{fig:more_res}) successfully capture the unique characteristics specified in the reference images.
This significantly enhances the controllability of the editing process, presenting considerable practical value.
In both qualitative and quantitative comparisons, \sysname{} consistently demonstrates superior performance in editing quality, visual fidelity, and user satisfaction when compared to existing methods.

Our contributions can be summarized as follows: 
\begin{itemize}
\item We present \sysname{}, a versatile 3D scene editing framework that allows the users to perform various editing operations 
(e.g. object insertion, object replacement, re-texturing, and stylization)
guided by not only the text prompt but also by a reference image.
\item We present a novel stepwise 2D personalization strategy, 
which features a localization loss in the scene personalization step and a separate novel content personalization step dedicated to the reference image based on LoRA,
to enable accurate location and appearance control.
\item We adopt 3D Gaussian splatting to represent scenes due to its rendering efficiency and, more importantly, its explicit point data structure, which is very suitable for precise local editing. 
\end{itemize}

\section{Related Works}

% 主要说一下基于Diffusion model 的图像编辑
\subsubsection*{Text-guided image generation and editing}
Text-to-image (T2I) diffusion models ~\cite{ramesh2022hierarchical, saharia2022photorealistic, rombach2022high}, trained on large-scale paired image-text datasets, 
have gained significant attention 
since they can generate diverse and high-quality images that match the complicated text prompt.
Instead of directly generating images from scratch, another popular and closely related task is to edit the given image according to the text prompt~\cite{meng2021sdedit,couairon2022diffedit,kawar2022imagic,hertz2022prompt,avrahami2022blended,brooks2022instructpix2pix}.

Another popular task is object/concept personalization, which aims at generating images for a specified object/concept defined in the given image collection.
Textual Inversion (TI)~\cite{gal2022image} 
optimizes special text token(s) in the text embedding space to represent the specified concept.
DreamBooth~\cite{ruiz2022dreambooth} fine-tunes the entire diffusion model with a class-specific prior preservation loss as regularization. 
In general, DreamBooth generates higher-quality images since it involves a larger amount of updated model parameters (i.e. the whole UNet model).
However, all the aforementioned methods do not support generating images containing multiple personalized objects simultaneously.

Custom Diffusion~\cite{kumari2023multi} extends the above task to generate multiple personalized \emph{concepts} in one image simultaneously.
Although separate special text tokens are assigned to each \emph{concept}, the UNet is updated by all \emph{concepts}, resulting in less satisfactory personalization results.
Furthermore, it lacks a localization mechanism to specify the interaction between two \emph{concepts} (Fig.~\ref{fig_custom}).
In contrast, we propose a stepwise 2D personalization strategy to learn the existing scene and the new content separately, achieving high-quality and faithful personalization results and being generalizable to sequential editing scenarios.

\subsubsection*{Radiance field-based 3D object/scene generation}

The success of T2I diffusion models has largely advanced the development of 3D object/scene generation.
One seminal contribution, DreamFusion~\cite{poole2022dreamfusion}, introduces score distillation sampling (SDS), which distills knowledge from a pre-trained 2D T2I model to \emph{optimize} a radiance field without the reliance on any 3D data.
Most of the subsequent works adopt such an optimization-based pipeline and make further progresses by introducing an extra refinement stage (e.g., Magic3D~\cite{lin2022magic3d} and DreamBooth3D~\cite{raj2023dreambooth3d}), or proposing more suitable SDS variants (e.g., VSD~\cite{wang2023prolificdreamer}), or using more powerful 3D representations~\cite{chen2023fantasia3d,yi2023gaussiandreamer,chen2023text}, such as GS~\cite{kerbl20233d}.

Furthermore, a body of research~\cite{deng2022nerdi, melas2023realfusion, tang2023make, qian2023magic123} endeavors to integrate reference images within the optimization framework. This integration is facilitated by various techniques, including the application of reconstruction loss, employment of predicted depth maps, and the execution of a fine-tuning process. 
Nevertheless, these methods are constrained to generate a single object from scratch and cannot edit existing 3D scenes.

\subsubsection*{Radiance field-based 3D object/scene editing}
Earlier works~\cite{wang2022clip,wang2023nerf} mainly focus on global style transformation of a given 3D scene, which takes text prompts or reference images as input and usually leverage a CLIP-based similarity measure~\cite{radford2021learning} during optimization.
Several studies enable local editing on generic scenes by 
utilizing 2D image manipulation techniques (e.g. inpainting)~\cite{liu2022nerf, kobayashi2022decomposing,bao2023sine} to obtain new training images to update the existing radiance field.
Some works adopt 3D modeling techniques (e.g. mesh deformation~\cite{yuan2022nerf, yang2022neumesh, xu2022deforming, wang2023mesh} or point clouds deformation ~\cite{wu2023nerf, chen2023neuraleditor}) to propagate the shape deformation to the underlying radiance field.
However, these methods require extensive user interactions.

Recently, text-driven radiance field editing methods have gained more and more attention for their editing flexibility and accessibility. 
For example, Instruct-NeRF2NeRF~\cite{haque2023instruct} and GenN2N~\cite{liu2024genn2n} employ an image-based diffusion model (InstructPix2Pix~\cite{brooks2022instructpix2pix}) to modify the rendered image by the users' instructions, and subsequently update the 3D radiance field with the modified image.
DreamEditor~\cite{zhuang2023dreameditor}, 3D Paintbrush~\cite{decatur20233d} and Vox-E~\cite{sella2023vox} enable better local editing by adopting explicit 3D representations (i.e. mesh and voxel), where the editing region is automatically determined by the 2D cross-attention maps.
GaussianEditor~\cite{chen2023gaussianeditor, fang2023gaussianeditor} adopts GS as the scene representation and incorporates 3D semantic segmentation~\cite{kirillov2023segment, cen2023segment} to facilitate efficient and precise scene editing.
However, these text-driven approaches lack precise control over the specified appearance and position of the editing results.

A concurrent work, CustomNeRF~\cite{he2023customize}, is most related to our task setting.
But CustomNeRF only supports the object replacement task, since it requires an object that can be detected by the segmentation tool~\cite{kirillov2023segany} existing in the implicit NeRF scene, as the editing target.
In contrast, we adopt explicit GS as the 3D representation which facilitates our method to perform more editing tasks (e.g., object insertion and stylization).

\section{Background}

\subsubsection*{3D Gaussian Splatting.}
3D Gaussian Splatting (GS)~\cite{kerbl20233d} quickly draws tremendous attention due to its high rendering quality and efficiency.
GS utilizes a set of point-like anisotropic Gaussians $g_i$ to represent the scene: $\mathcal{G} = \{g_1, g_2, ... , g_N\}$.
Each $g_i$ contains a series of optimizable attributes, including center position $\mu \in \mathbb{R}^3$, opacity $\alpha \in \mathbb{R}^1$, 3D covariance matrix $\Sigma$, and color $c$.
The differentiable splatting rendering process is outlined as follows:
\begin{equation}
C =  \sum_{i\in \mathcal{N}} c_i \sigma_i \prod_{ i-1}^{j=1}  (1-\sigma_j), \quad \sigma_i= \alpha_i G(x) = \alpha_i e^{-\frac{1}{2} (x)^{T} \Sigma^{-1}(x) },
\end{equation}
where $j$ indexes the Gaussians in front of $g_i$ according to their distances to the optical center in ascending order,
$\mathcal{N}$ is the number of Gaussians that have contributed to the ray,
and $c_i$, $ \alpha_i $, and $x_i $ represent the color, density, and distance to the center point of the $ i $-th Gaussian, respectively.

\subsubsection*{Optimizing Radiance Fields with SDS Loss.}
Score distillation sampling (SDS) ~\cite{poole2022dreamfusion} optimizes a radiance field by distilling the priors from a Text-to-Image (T2I) diffusion model for 3D generation.
The pre-trained diffusion model $\phi$ is used to predict the added noise given a noised image $\hat{I}_{t}$ and its text condition $y$. 
\begin{equation}
  \nabla_{\theta}\mathcal{L}_{SDS}(\phi, \hat{I}=f(\theta))=\mathbb{E}_{\epsilon,t}\bigg[w(t)(\epsilon_{\phi}(\hat{I}_{t};y,t)-\epsilon)\frac{\partial \hat{I}}{\partial \theta } \bigg],
\end{equation}
where $\theta$ denotes the parameters of the radiance field, 
$f(\cdot)$ is the differentiable image formation process,
and $w(t)$ is a predefined weighting function derived from noise level $t$.

\begin{figure*}
\setlength{\abovecaptionskip}{0.2cm}
\includegraphics[width=1.0\textwidth]{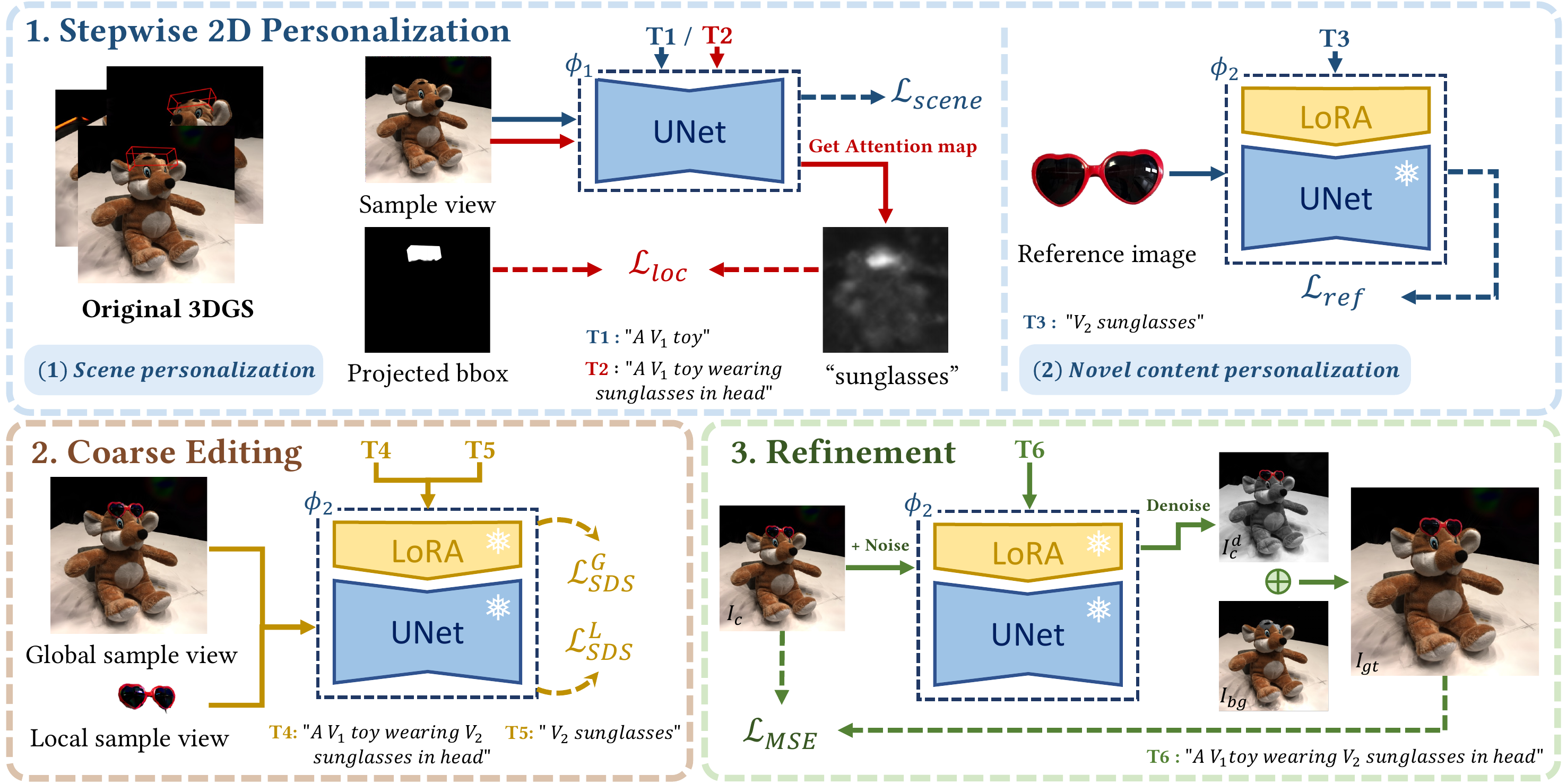}
\caption{
\textbf{Method overview.}
\sysname{} optimizes a 3D scene that is represented as 3D Gaussian splatting (GS) to conform with a given hybrid text-image prompt.
The editing process includes three stages: 
1) a stepwise 2D personalization strategy, which features a localization loss in the scene personalization step and 
a separate novel content personalization step dedicated to the reference image based on LoRA (Sec.~\ref{subsec:learning});
2) a coarse editing stage using SDS (Sec.~\ref{subsec:coarse});
and 3) a pixel-level texture refinement stage, utilizing carefully generated pseudo-GT image from both the rendered image $I_c$ and the denoised image $I_c^{d}$ (Sec.~\ref{subsec:refinement}).
}
\label{fig:overview}
\end{figure*}

\section{Method}

Given posed images (i.e., images and their associated camera parameters estimated by COLMAP~\cite{schoenberger2016sfm}) of the target scene, 
our goal is to enable more accurate editing following a hybrid text-image prompt within a user-specified 3D bounding box.
We choose 3D Gaussian splatting (GS)~\cite{kerbl20233d} to represent the 3D scene since GS is an explicit and highly flexible 3D representation method, which is beneficial for the following editing operations, especially local editing.

As shown in Fig.~\ref{fig:overview}, \sysname{} contains three major steps, including 
1) a stepwise 2D personalization of the existing scene and the novel content (Sec.~\ref{subsec:learning}), 
2) a coarse 3D editing stage using score distillation sampling (SDS)~\cite{poole2022dreamfusion} (Sec.~\ref{subsec:coarse}), 
and 3) a pixel-level refinement of the 3D scene (Sec.~\ref{subsec:refinement}).

\subsection{Stepwise 2D Personalization}
\label{subsec:learning}

In general, our stepwise personalization of the pre-trained T2I model (i.e., Stable Diffusion (SD)~\cite{rombach2022high}) is based on DreamBooth~\cite{ruiz2022dreambooth}, but with two significant modifications. These changes are essential to personalize both the existing scene and the novel content in the reference image. 
First, in the 2D personalization of the existing scene, we propose an attention-based localization loss to enforce the interaction between the existing and the novel content specified by the provided 3D bounding box (e.g., sunglasses on the forehead, see Fig.~\ref{fig_ablation_learning_1}).
Note that the reference image is not involved in this step.
Second, in the 2D personalization of the novel content, we introduce LoRA layers to better capture the unique characteristics of the specified item in the reference image.

\vspace{0.2cm}
\subsubsection*{2D personalization of the existing scene.}
We first personalize the SD to the given scene to facilitate various types of editing of the scene afterward.
Specifically, the initial text prompt (e.g. "a toy") is obtained using an image captioning model, BLIP-2~\cite{li2023blip}.
To enhance the specificity of the scene, we add a special token $V_1$ in front of the noun describing the scene, resulting in a scene-specific text prompt (e.g., ``a $V_1$ toy'') as in~\cite{zhuang2023dreameditor}.
The UNet $\epsilon_{\phi}$ of the T2I model is fine-tuned with the reconstruction loss and the prior preservation loss~\cite{ruiz2022dreambooth}.
The input of the reconstruction training includes the scene-specific text and a rendered image of the 3D scene from a random view.
The input of the prior preservation training includes the initial text and a random image generated by SD using the initial text as input (omitted in Fig.~\ref{fig:overview} to reduce clutter).
The above losses are computed as follows:
\begin{equation}
\begin{split}
\mathcal{L}_{scene}  = & \mathbb{E}_{z,y,\epsilon,t}|| \epsilon_{\phi_1}(z_t,t,p,y) -\epsilon||_{2}^{2} +  \\
 &\mathbb{E}_{z^*,y^*,\epsilon,t^*}|| \epsilon_{\phi_1}(z^*_t,t^*,p^*, y^*) - \epsilon||_{2}^{2}
\end{split}
\label{equ:loss_scene}
\end{equation}
where $y$ denotes the text prompt, $t$ the timestep, $z_{t}$ the noised latent code at $t$-th timestep extracted from the input scene image, and $p$ the camera pose.
Superscript $*$ denotes the corresponding variables used in prior preservation training.
Note that we add an additional camera pose $p$ to the condition embeddings in the network to have a better viewpoint control of the generated images from the SD, facilitating the subsequent SDS-based 3D scene optimization.
Since randomly generated images for prior preservation training do not have a meaningful ``scene pose'', we assign a fixed camera pose $p^*=I_4$ that will never be used for rendering.

To encourage accurate localization of the target object,
we introduce an attention-based localization loss (Fig.~\ref{fig:overview}) during the 2D scene personalization to encourage the SD to generate images containing the required scene-object interaction.
This step is particularly important if the target object is specified at a rarely seen location (e.g., sunglasses on the forehead, see Fig.~\ref{fig_ablation_learning_1}).
The actual location of the target object generated by SD is extracted from the cross-attention map $A_{t}$ of the object keyword (e.g., ``sunglasses'') following~\cite{hertz2022prompt}.
The wanted location of the target object (i.e., GT editing region) is obtained by projecting the provided 3D bounding box to the image plane.
The loss between the actual and the wanted location is defined as:
\begin{equation}
\mathcal{L}_{loc} = (1 - \underset{s\in \mathcal{S}}{max}(A_t^{s})) + \lambda \sum_{s\in \bar{\mathcal{S}}}^{} || A_t^{s} ||_{2}^{2}
\label{equ:loss_att}
\end{equation}
where, $\lambda$ is a weight to balance two terms, $\mathcal{S}$ the GT editing mask region (projection of the 3D bounding box $\mathcal{B}$) and $\bar{\mathcal{S}}$ the otherwise. 
Intuitively, this loss encourages a high probability 
inside the editing area and penalizes the presence of the target object outside the editing area.
As demonstrated in our ablation study (Fig.~\ref{fig_ablation_learning_1}), this loss is crucial for ensuring precise editing within the specified region.

\subsubsection*{2D personalization of the novel content.}
We introduce a dedicated personalization step using LoRA~\cite{hu2021lora} (with the UNet fixed) to better capture the unique characteristics contained in the reference image.
This step is essential to reduce the negative influence (e.g. concept forgetting~\cite{kumari2023multi}) when learning (personalizing) multiple concepts, 
resulting in a better representation of both the scene and the novel content.
Specifically, we train the additional LoRA layers inserted to the previously personalized and fixed T2I model $\epsilon_{\phi^{*}}$.
Similar to the last step, we obtain the initial text prompt using BLIP-2 model and insert a special token $V_2$ into it, 
yielding an object-specific text prompt $y^r$ of the reference object (e.g. ``$V_2$ sunglasses'').
The new LoRA layers are trained with the following loss function:
\begin{equation}
\mathcal{L}_{ref} = \mathbb{E}_{z^{r},y^{r},\epsilon,t}|| \epsilon_{\phi_2}(z_t^{r},t,p^*,y^{r}) - \epsilon||_{2}^{2}
\label{equ:loss_subject}
\end{equation}
After training, the content of the scene and the reference image are stored in UNet and added LoRA layers, respectively,
resulting in largely reduced mutual interference.

\subsection{Coarse Editing via SDS Loss}
\label{subsec:coarse}

We optimize the selected Gaussians $\mathcal{G^{\mathcal{B}}} \in \mathcal{B}$ (i.e., those inside the bounding box $\mathcal{B}$) with SDS loss from the personalized T2I diffusion model $\epsilon_{\phi_2}$.
Specifically, we input randomly rendered images $\hat{I}$ using sampled camera poses $p$ and the text prompt $y^{G}$ into the T2I model $\epsilon_{\phi_2}$, and calculate the global scene SDS Loss as follows: 
\begin{equation}
\nabla_{\mathcal{G} }\mathcal{L}_{SDS}^G(\phi_2, f(\mathcal{G} )) = \mathbb{E}_{\epsilon,t}\bigg[w(t)(\epsilon_{\phi_2}(z_{t};t,p,y^{G})-\epsilon)\frac{\partial z}{\overset{}{\partial} \hat{I} } \frac{\partial \hat{I}}{\overset{}{\partial}\mathcal{G}} \bigg]
\label{equ:sds_global}
\end{equation}
where $y^{G}$ is the text prompt including special tokens $V_1, V_2$ and describes our wanted result,
$f(\cdot)$ the GS rendering algorithm.

It is noteworthy that the selection and update criteria of the Gaussians $\mathcal{G^{\mathcal{B}}}$ to be optimized are slightly different for different types of editing tasks.
For object insertion, we duplicate all the Gaussians inside the bounding box and exclusively optimize all the attributes of these new Gaussians.
For object replacement and re-texturing, all the Gaussians inside the bounding box will be updated.
For stylization, optimization is applied to all the Gaussians in the scene.
Note that we only update the colors (i.e., the spherical harmonic coefficients) for re-texturing instead of updating all the attributes.

Since the foreground and background of a GS-based scene are readily separable given the bounding box $\mathcal{G^{\mathcal{B}}}$, we introduce another local SDS loss for object-centric editing (e.g., object insertion/replacement) to reduce artifacts as follows:
\begin{equation}
\nabla_{\mathcal{G^{\mathcal{B}}} }\mathcal{L}_{SDS}^L(\phi_2, f( \mathcal{G^{\mathcal{B}}} )) = \mathbb{E}_{\epsilon,t}\bigg[w(t)(\epsilon_{\phi_2}(z_{t};t,p,y^L)-\epsilon) \small{\frac{\partial z}{\overset{}{\partial} \hat{I} } \frac{\partial \hat{I}}{\overset{}{\partial}\mathcal{G^{\mathcal{B}}}} } \bigg]
\label{equ:sds_local}
\end{equation}
where $y^L$ is the text prompt including the special tokens $V_2$ and only describes our wanted new object,
$\hat{I}$ the rendered images containing only the foreground object.

We employ $\mathcal{L}_{SDS}^G$ and $\mathcal{L}_{SDS}^L$ with weight $\gamma$ to optimize $\mathcal{G^{\mathcal{B}}}$:
\begin{equation}
\mathcal{L_{\mathcal{SDS}}}= \gamma \mathcal{L}_{SDS}^G + (1-\gamma)\mathcal{L}_{SDS}^L
\label{equ:sds}
\end{equation}

\subsection{Pixel-Level Image Refinement}
\label{subsec:refinement}
In this stage, we introduce a pixel-level reconstruction loss to effectively enhance the quality of the editing results, 
since the 3D results directly optimized with SDS loss usually contain artifacts
(e.g. green noise on the glasses' frame, needle-like noise on the hair in Fig.~\ref{fig:corase_res}).

The core of this stage is to create a pseudo-GT image $I_{gt}$ to supervise the rendered image $I_c$ from the coarse GS.
Firstly, we follow SDEdit~\cite{meng2021sdedit} to add noise on the rendered image $I_c$ to obtain the noised image $I_c^{d}$ and 
then utilized the personalized T2I model $\epsilon_{\phi_2}$ as a denoising network and obtain the denoised image  $I_c^{d}$.
The denoising process effectively reduces the artifacts in $I_c$ (see Fig.~
D.1 in the supplementary), but also alters the background image.
Secondly,
we obtain the binary instance mask $M^{inst}$ of the edited object/part
by rendering only the editable Gaussians $\mathcal{G}^\mathcal{B}$ and thresholding its opacity mask.
Then, we render a background image $I_{bg}$ with only the fixed Gaussians.
Finally, the pseudo-GT image $I_{gt}$ is obtained as:
\begin{equation}
I_{gt} = M^{inst} \odot I_c^{d} + (1-M^{inst}) \odot I_{bg}
\end{equation}
This process ensures that the background image is clean and the same as the original scene while the foreground editable region is enhanced by the T2I model $\epsilon_{\phi_2}$.
Using this pseudo-GT image as pixel-level supervision effectively enhances the resultant texture and reduces floaters (Fig.~\ref{fig:corase_res}).
MSE loss is applied between the rendered image $I_c$ and the created pseudo-GT image $I_{gt}$.
A flowchart depicting the complete preparation of $I_{gt}$ is included in the supplementary.

To maintain better coverage, the rendering camera poses cover all elevation and azimuth angles with an interval of $30^\circ$ within a predefined range.
To maintain better view-consistency of the denoised images, we set a small noise level ($t_{0}=0.05$, .i.e, ``intermediate time" in SDEdit).
Using such a small noise level effectively enhances fine texture details, removes small artifacts, and does not introduce significant shape and appearance change,  maintaining better view consistency for the target editing region.

\section{Experiments}

\begin{figure*}
\setlength{\abovecaptionskip}{0.2cm}
\includegraphics[width=1.0\textwidth]{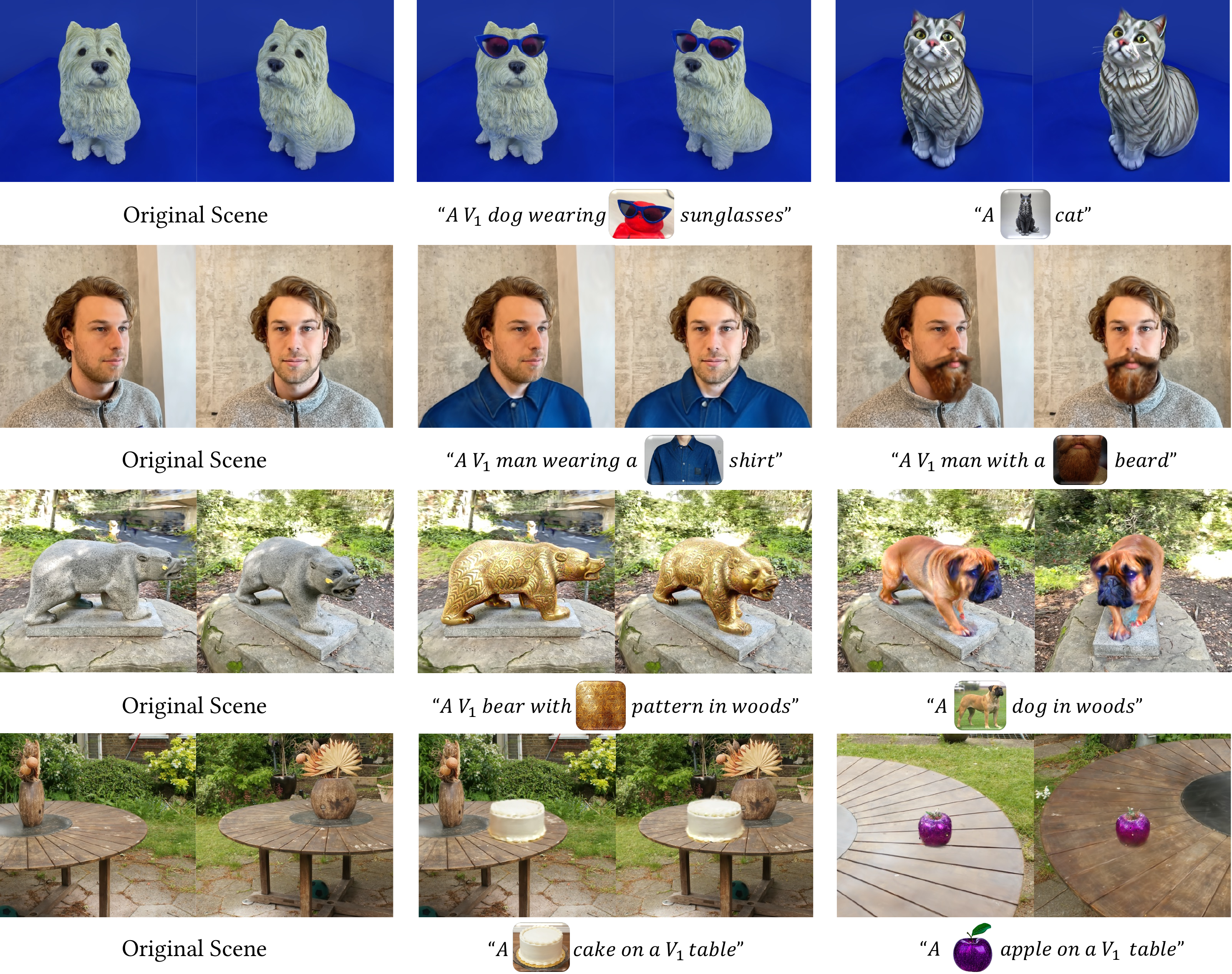}
\caption{
Editing results of the proposed \sysname{}.
Images in the text prompts denote their associated \emph{rare tokens}, which are fixed without optimization.
}
\label{fig:more_res}
\end{figure*}

\subsection{Experimental Setup}

\subsubsection*{Implementation Details}
We use the official code to train the original scene GS, with the default hyper-parameters.
In the stepwise 2D personalization stage, the scene personalization step consists of 1k iterations, while the novel content personalization contains 500. We set $\lambda=0.1$ in $\mathcal{L}_{loc}$. 
In the coarse editing stage, we adopt the sampling strategy of views from ~\cite{zhuang2023dreameditor}.  The size of the rendered images is 512$\times$512. 
Owing to the different complexity of the editing task, this stage requires optimizing for 1K$\sim$5K iterations, consuming approximately 5$\sim$25 minutes.
The refinement stage takes 3K iterations with the supervision of the generated $I_{gt}$, concluding in less than 3 minutes.
More implementation details can be found in the supplementary.

\subsubsection*{Dataset}
To comprehensively evaluate our method, 
We select six representative scenes with different levels of complexity
following previous works~\cite{zhuang2023dreameditor,haque2023instruct,chen2023gaussianeditor}. 
These scenes include objects in simple backgrounds, human faces, and complex outdoor scenes.
For each editing, a hybrid prompt, consisting of text and a reference image obtained from the Internet, is employed to guide the editing. 
Additionally, we manually set a 3D bounding box to define the editing region.

\subsubsection*{Baselines}
Due to the lack of dedicated image-based editing baselines, we compare with two state-of-the-art text-based radiance field editing methods, including Instruct-NeRF2NeRF (``I-N2N'')~\cite{haque2023instruct} and DreamEditor~\cite{zhuang2023dreameditor}.
I-N2N utilizes Instruct-pix2pix~\cite{brooks2022instructpix2pix} to update the rendered multi-view images according to special text instructions.
DreamEditor adopts a mesh-based representation and includes an attention-based localization operation to support local editing.
For a fair comparison, we replace its automatic localization with a more accurate manual selection.
See our supplementary for more implementation details.

\begin{figure*}
\includegraphics[width=1.0\textwidth]{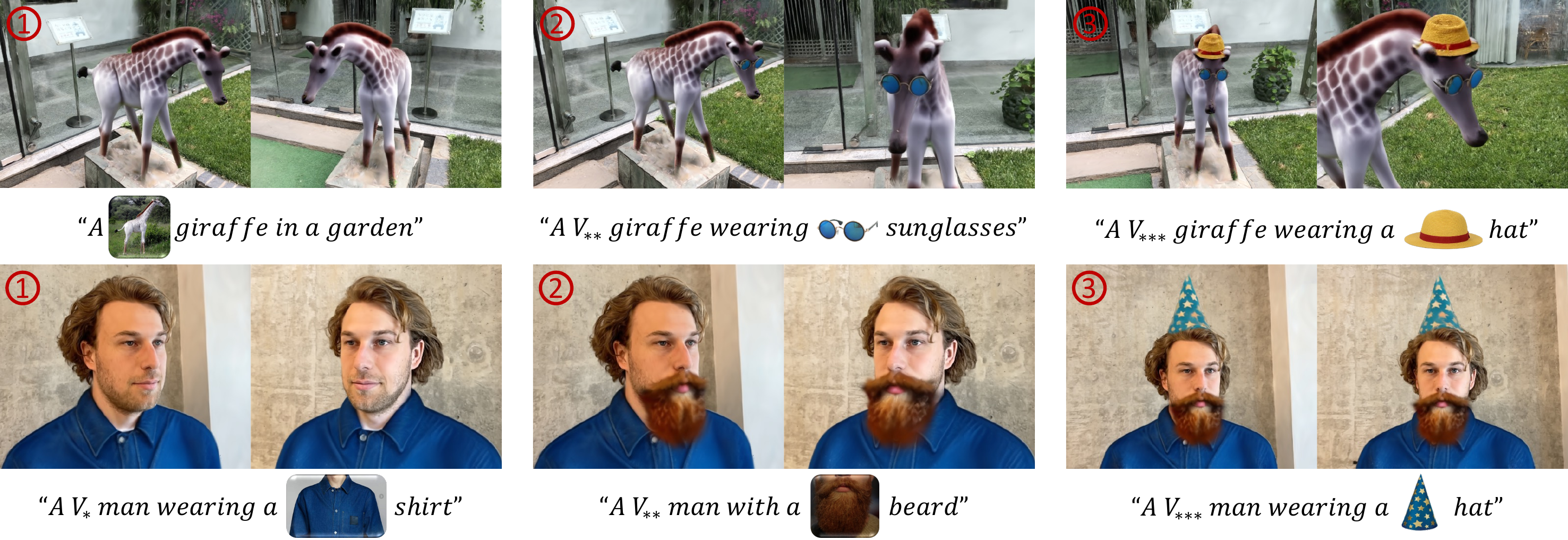}
\caption{
Sequential editing results. 
We show two rendered images of the 3D scene after every editing step, indicated by the number in the top-left corner.
$V_*$, $V_{**}$, and $V_{***}$ represent the special tokens of the scene in different sequences of editing.}
\label{fig:sequence_res}
\end{figure*}

\begin{figure*}
\includegraphics[width=1.0\textwidth, height=0.285\textheight]{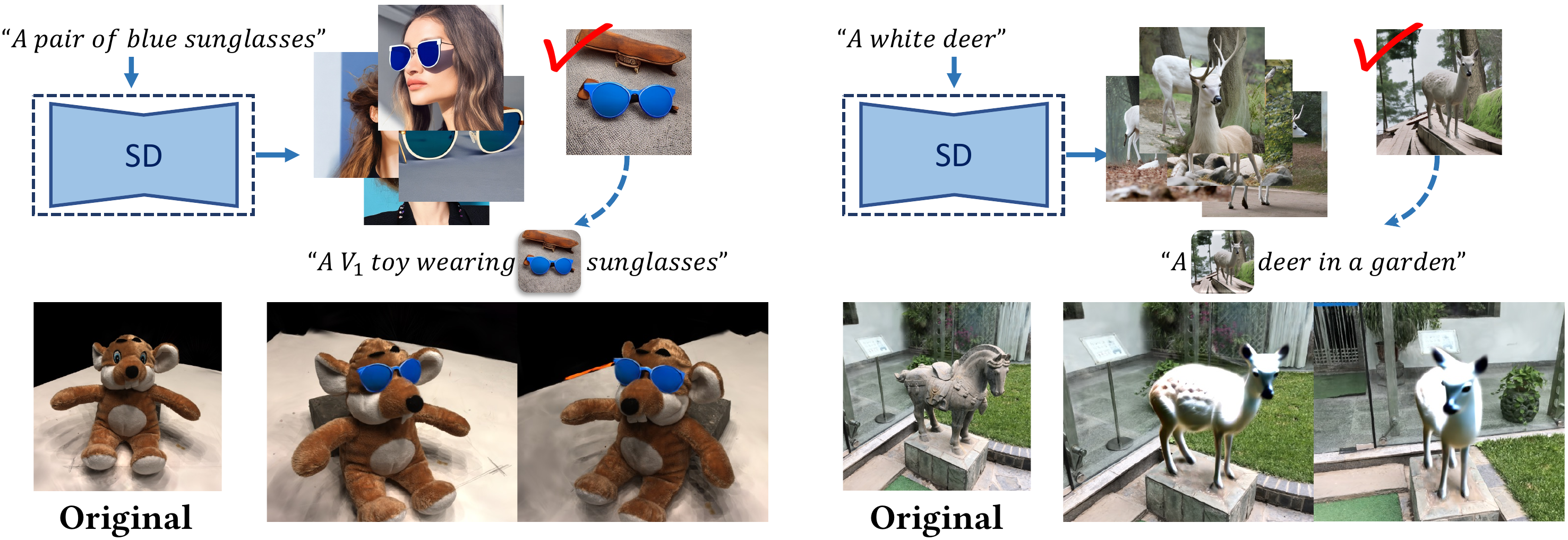}
\caption{Results of using a generated image as the reference. We first generate several candidate images by the diffusion model using text prompts, then we choose one as the reference image for editing. }
\label{fig:text_res}
\end{figure*}

\subsubsection*{Evaluation Criteria}
For quantitative evaluation, we adopt CLIP Text-Image directional similarity following~\cite{haque2023instruct, zhuang2023dreameditor} to assess the alignment of the editing outcomes with the given text prompt.
To evaluate image-image alignment (between the edited scene and the reference image), we follow~\cite{he2023customize} to calculate the average DINO similarity~\cite{oquab2023dinov2} between the reference image and the rendered multi-view images of the edited 3D scene.
Detailed information about these calculations is available in the supplementary.

Additionally, we conduct a user study and ask the participants (50 in total) to evaluate the results of different methods from two aspects (overall ``Quality'', and ``Alignment'' to the reference image).
The user study includes 10 questions, each containing the edited results of the two baselines and ours rendered into rotating videos in random order (see our supplementary).
The 10 questions have covered various scenes and editing types to better compare the methods under different scenarios.

\subsection{Visual Results of \sysname{}}

In Fig.\ref{fig:teaser} and Fig.~\ref{fig:more_res}, we present qualitative results of \sysname{}.
Video demonstrations are included in the supplementary.
Experiments on diverse 3D scenes demonstrate that \sysname{} effectively executes various editing tasks, including re-texturing, object insertion, object replacement, and stylization, achieving both high-quality results and strictly following the provided text prompt and reference image.

\subsubsection*{Keeping unique characteristics specified by the reference image}
One of the most distinguishable differences between \sysname{} and previous methods is that \sysname{} also supports an image prompt, which offers more accurate control and makes it more user-friendly in real applications.
Results in Fig.~\ref{fig:teaser}\&\ref{fig:more_res} demonstrate high consistency between the updated 3D scene and the reference image
(e.g. the \emph{styles} of the sunglasses; the \emph{white} giraffe; the \emph{virtual ghost} horse; the joker make-up appeared in movie \emph{The Dark Knight}).
Moreover, as depicted in the bottom of Fig.~\ref{fig:teaser}, our method can also perform global scene editing, 
such as transferring the entire scene in the \emph{Modigliani} style of the reference image.

\begin{figure*}
\includegraphics[width=1\textwidth]{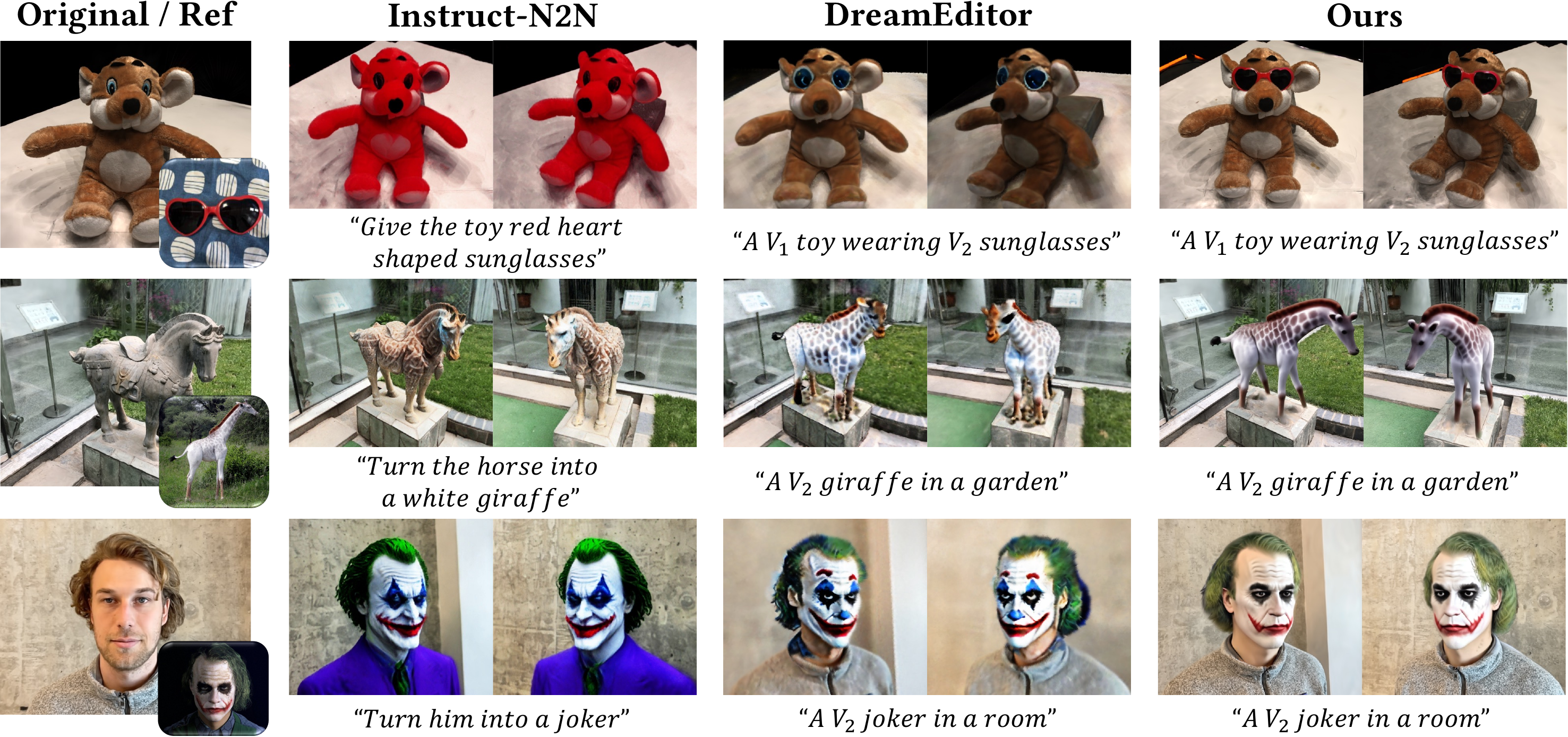}
\caption{
Visual comparisons between different methods.
Our method produces obviously higher-quality results and \emph{accurately} follows the reference image input (bottom-right corner in column 1).
% 
% The provided image prompts of \sysname{} are shown at the corner in column 1.
\nerf2nerf{} sometimes misunderstands (row 1) or overlooks (row 2) the keywords.
%, or cannot generate a specified appearance in limited experiments(row 3).
DreamEditor faces difficulty in making obvious shape changes (row 2).
Both of them do not support image prompts to specify detailed appearance/style, producing less controlled results.
% In contrast, our method produces obviously higher-quality results and closely follows the reference image input (bottom-right corner in column 1).
% Also refer to user study result in Tab.~\ref{tab:Quantitative}.
}
% Visual results of our method compared with two baselines on three different scenes. 
% $\mathbf{Ref} $ denotes the reference image.
% The results of our \sysname{} distinctly demonstrate superior texture and geometry quality, alongside the highest consistency with the reference image.
\label{fig_visual_compare}
\end{figure*}

\subsubsection*{Sequential editing}
\sysname{} can sequentially edit the initial scene multiple times thanks to the local update of the GS and the stepwise 2D personalization strategy, which effectively reduces the interference between the existing scene and the novel content.
Results in Fig.\ref{fig:sequence_res} demonstrate the sequential editing capability.
There is no observable quality degradation after multiple times of editing and no interference between different editing operations.

\subsubsection*{Using generated image as the reference}
In the absence of the reference image,
we can generate multiple candidates from a T2I model and let the user choose a satisfactory one.
This interaction offers the user more control and makes the final result more predictable.
Fig.~\ref{fig:text_res} shows some examples.

\begin{table}[t]
\caption{Quantitative comparisons. CLIP$_{dir}$ denotes the CLIP Text-Image directional similarity. DINO$_{sim}$ is the DINO similarity.
}
\label{tab:Quantitative}
\renewcommand\tabcolsep{4pt}
\small
\begin{tabular}{lcccc}
\toprule
Method  & CLIP$_{dir} \uparrow$  & DINO$_{sim} \uparrow$  & Vote$_{quality}$ & Vote$_{alignment}$ \\
\midrule
\nerf2nerf{} & 8.3 & 36.4 & 21.6\% & 8.8\% \\
DreamEditor &  11.4  & 36.8 & 7.6\%  & 10.0\%\\
Ours & \textbf{15.5} & \textbf{39.5} & \textbf{70.8\%} & \textbf{81.2\%} \\
\bottomrule
\end{tabular}
\end{table}

\subsection{Comparisons with State-of-the-Art Methods}

\subsubsection*{Qualitative comparisons.}

Fig.\ref{fig_visual_compare} shows visual comparisons between our method and the baselines.
Since both baselines do not support image prompts as input, 
they generate an uncontrolled (probably the most common) item belonging to the object category.
In contrast, our results consistently maintain the unique characteristics specified in the reference images (i.e., the \emph{heart-shaped} sunglasses; the \emph{white} giraffe; the joker from the movie \emph{The Dark Knight}).

Moreover, \nerf2nerf{} sometimes misunderstands (row 1) or overlooks (row 2) the keywords, or cannot generate a specified appearance in limited experiments (row 3), 
probably due to limited supported instructions in Instruct-Pix2Pix.
DreamEditor also faces difficulty if the user wants to add a specified sunglasses item (row 1).
Additionally, it is difficult for DreamEditor to make obvious shape changes (row 2) to the existing object due to its adoption of a less flexible mesh-based representation (i.e., NeuMesh).

\begin{figure}
\includegraphics[width=\columnwidth]{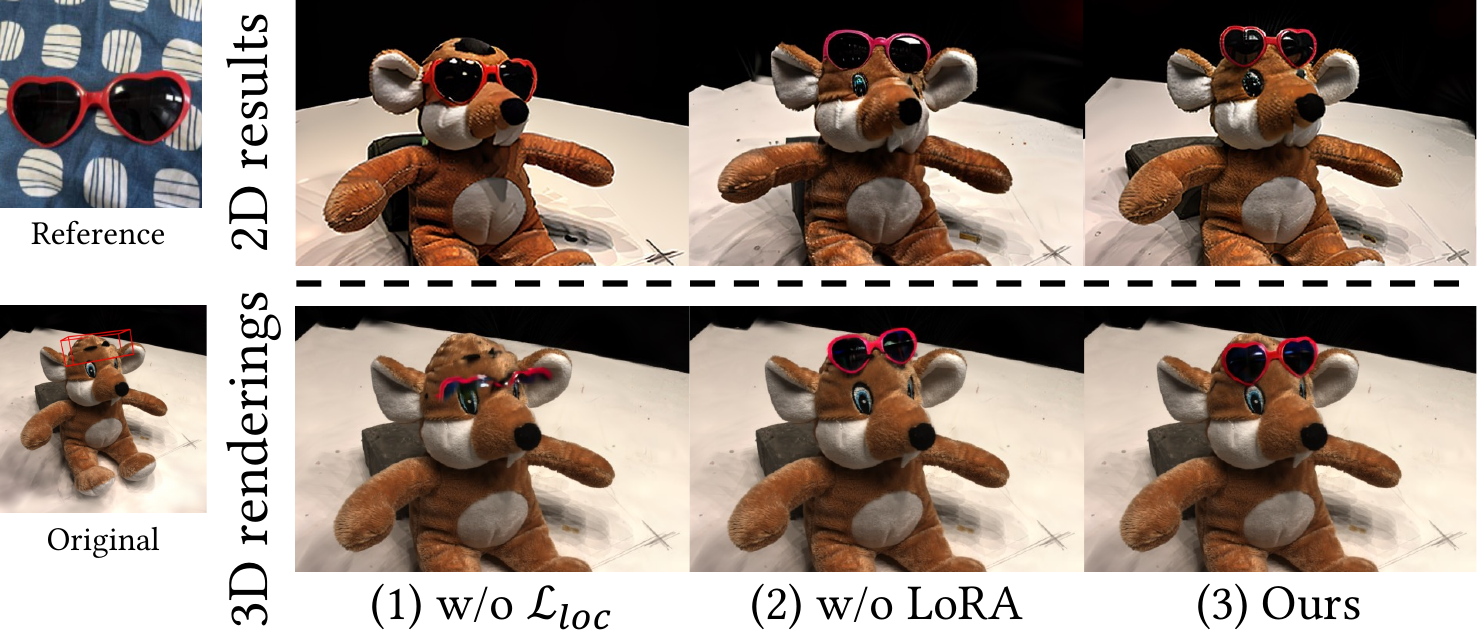}
\caption{
Ablation study on the components proposed in stepwise 2D personalization.
We compare the generated images of the personalized T2I model (top row) and the rendered images of the updated 3D scene (bottom row).
Removing the localization loss $\mathcal{L}_{loc}$ fails to place the new object in the specified place.
Removing the separate LoRA layers dedicated for the personalization of the reference image produces less similar results (heart-shaped vs. regular round shape).
% The top row illustrates the image output of the T2I diffusion model employing three variants of personalization strategies.
% The bottom row shows the corresponding editing results within the 3D scene.
% The $\mathcal{L}_{loc}$ ensures targeted editing within the specified region (i.e. on the head). 
% Without $\mathcal{L}_{loc}$, sunglasses tend to appear on the eyes, constituting a more conventional combination.
% }
% The introduction of LoRA Layers enhances the ability to accurately represent the appearance and shape of the reference object.
}
\label{fig_ablation_learning_1}
\end{figure}

\begin{table}[t]
\caption{
Quantitative evaluation of the 3D renderings in the ablation study on the components proposed in stepwise 2D personalization (Fig.~\ref{fig_ablation_learning_1}).
}
\label{tab:fig_ablation_learning_1}
\small
\renewcommand\tabcolsep{15pt}
\begin{tabular}{lccc}
\toprule
  & w/o $\mathcal{L}_{loc}$  & w/o LoRA & Ours  \\
\midrule
CLIP$_{dir}$ $\uparrow$ & 4.4 & 24.0 & \textbf{25.4} \\
DINO$_{sim}$ $\uparrow$ &  18.3  & 28.0  & \textbf{28.8}  \\
\bottomrule
\end{tabular}
\end{table}

\begin{figure*}
\includegraphics[ width=1.0\textwidth]{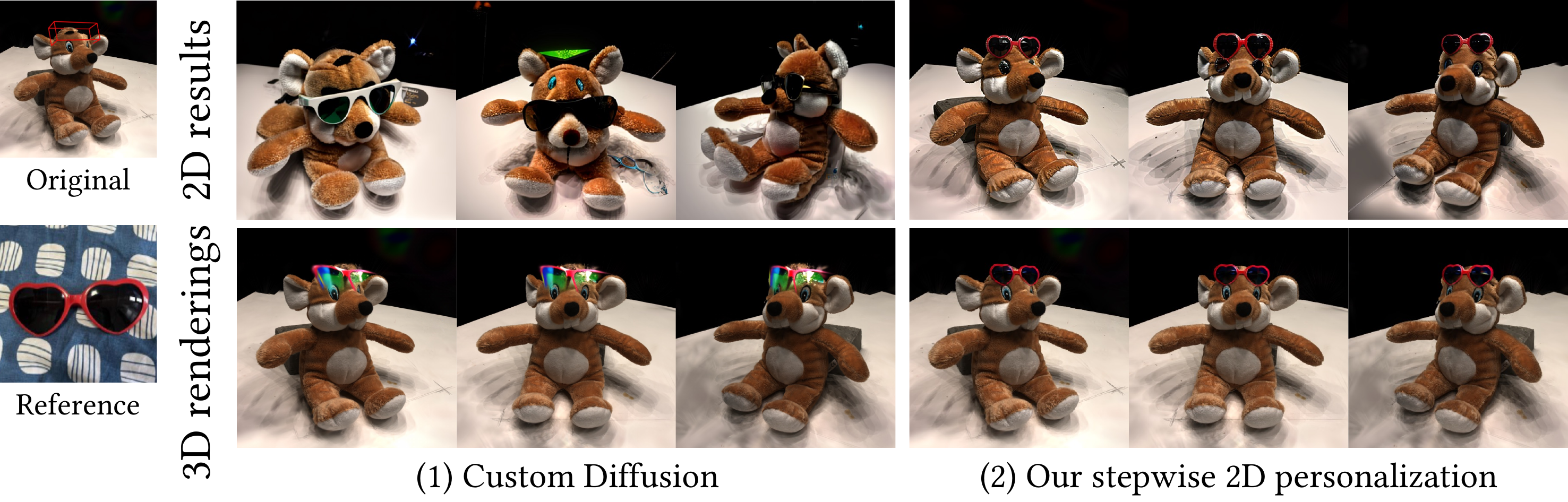}
\caption{
Comparison of different 2D personalization methods. 
Generated images of the T2I models after personalization (top) and the final updated 3D scene (bottom) are presented.
\emph{Text prompt}: ``A $V_1$ toy wearing $V_2$ sunglasses on the forehead"
}
\label{fig_custom}
\end{figure*}

\begin{figure}
\includegraphics[width=\columnwidth]{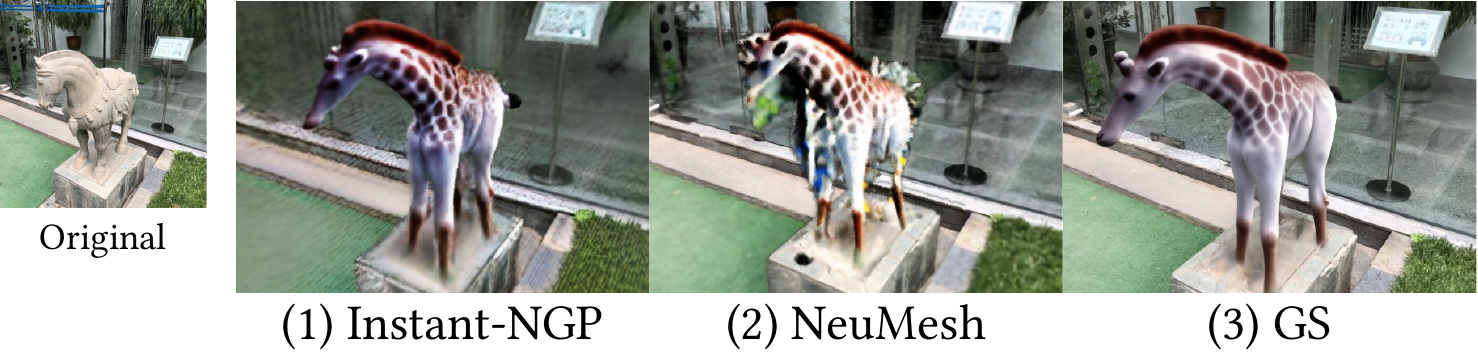}
\caption{
% \dk{
Ablation study on different 3D representations to show the advantage of GS for this task.
Using Instant-NGP results in a changed background while using NeuMesh cannot produce large enough shape deformation.
In contrast, using \emph{explicit} and \emph{flexible} GS obtains the best foreground editing result while keeping the background unchanged.
% }
% 
% Guided by the identical fine-tuned diffusion model, compared with Instant-NGP and NeuMesh, 3DGS generates results with superior texture geometry quality while maintaining the background unchanged.
}
\label{fig_ablation_learning_2}
\end{figure}
\begin{table}[t]
\caption{
Quantitative evaluation of the 3D renderings in the ablation study on different 3D representations (Fig.~\ref{fig_ablation_learning_2}).
}
\label{tab:ablation_3D_repr}
\small
\renewcommand\tabcolsep{15pt}
\begin{tabular}{lccc}
\toprule
  & Instant-NGP  & NeuMesh & Ours  \\
\midrule
CLIP$_{dir}$ $\uparrow$ & 17.8 & 18.4 & \textbf{18.7} \\
DINO$_{sim}$ $\uparrow$ &  31.8  & 27.0  & \textbf{33.5}  \\
\bottomrule
\end{tabular}
\end{table}

\begin{figure}
\includegraphics[width=\columnwidth]{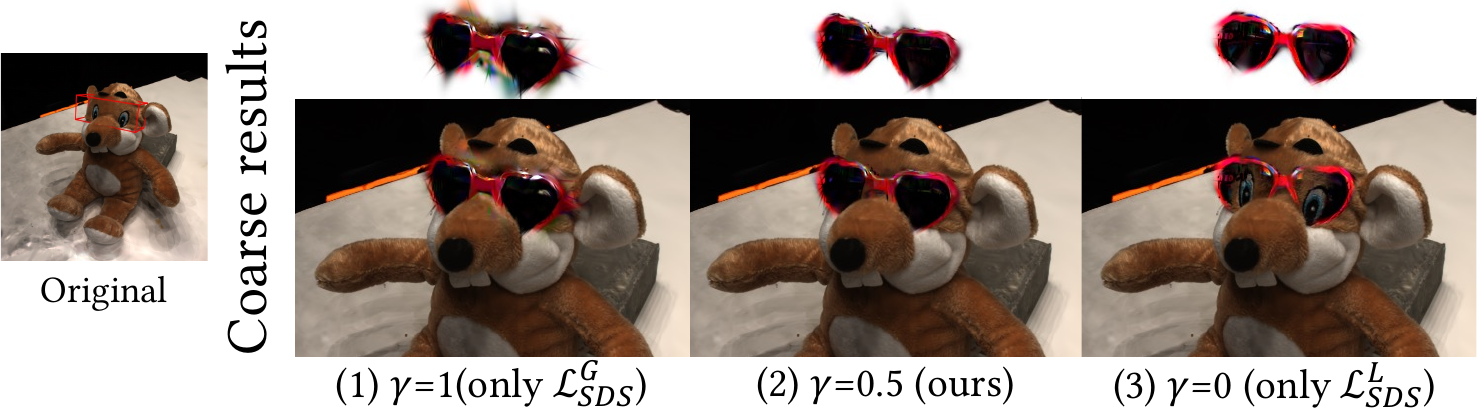}
\caption{
Ablation study on the influence of global and local SDS (Eq.~\ref{equ:sds}) in the coarse stage.
% both the global and local SDS loss are necessary and our solution achieves the best result.
% 
% Using global SDS loss $\mathcal{L}^{G}_{SDS}$ improves
The top row shows the rendering of the editable Gaussians $\mathcal{G^{\mathcal{B}}}$.
Only using global SDS $\mathcal{L}^{G}_{SDS}$ produces low-quality foreground object/part, while only using local SDS $\mathcal{L}^{L}_{SDS}$ produces unnatural foreground when composited with the existing scene (e.g., color, placement).
% Specifically, only using global SDS loss $\mathcal{L}^{G}_{SDS}$ results in obvious artifacts in the editable region.
% Only using local SDS loss $\mathcal{L}^{L}_{SDS}$ results in inaccurate placement of the object and unnatural color discrepancy between the background and the novel content since the context information is missing during optimization.
% \zjy{
% Coarse editing results optimized with different $\gamma$.
% We demonstrate the rendering images of the edited Gaussian $\mathcal{G^{\mathcal{B}}}$ (top row) and all Gaussians in the scene $\mathcal{G}$ (bottom row). 
% Combining $\mathcal{L}^{G}_{SDS}$ and $\mathcal{L}^{L}_{SDS}$ with $\gamma=0.5$ effectively reduces the noise Gaussian while correctly placing the sunglasses (Fig.\ref{fig_ablation_learning_3}(2)).
% }
}
\label{fig_ablation_learning_3}
\end{figure}
\begin{table}[t]
\caption{
Quantitative evaluation of the 3D renderings in the ablation study on different $\gamma$ in coarse editing (Fig.~\ref{fig_ablation_learning_3}).
}
\small
\label{tab:ablation_3D_repr}
\renewcommand\tabcolsep{18pt}
\begin{tabular}{lccc}
\toprule
  & $\gamma=1$  & $\gamma=0.5$ & $\gamma=0$  \\
\midrule
CLIP$_{dir}$ $\uparrow$ & 9.0 & \textbf{18.1} & 12.3 \\
DINO$_{sim}$ $\uparrow$ &  29.0  & \textbf{29.5}  & 27.5  \\
\bottomrule
\end{tabular}
\end{table}

\subsubsection*{Quantitative comparisons.}
Tab.~\ref{tab:Quantitative} shows the results of the CLIP Text-Image directional similarity and DINO similarity. 
The results clearly demonstrate the superiority of our method in both metrics, 
suggesting that the appearance generated by our method aligns better with both the text prompt and the image prompt.
A similar conclusion has been drawn according to the user study.
Our results surpass the baselines with a substantial margin on both the \emph{quality} evaluation ($70.8\%$ votes) and the \emph{alignment} evaluation ($81.2\%$ votes).

\subsection{Ablation Study}
\label{section:AS}

\subsubsection*{Ablation studies on the stepwise 2D personalization}

We conduct ablative experiments in Fig.\ref{fig_ablation_learning_1} and Tab.~\ref{tab:fig_ablation_learning_1} to demonstrate the benefit of using $\mathcal{L}_{loc}$ and LoRA Layers in the stepwise 2d personalization.
Without $\mathcal{L}_{loc}$, the fine-tuned T2I model fails to place the sunglasses in the specified region (i.e. on the forehead) due to the bias present in the training data of the original T2I model.
Introducing dedicated LoRA layers to personalize the unique features in the reference image results in more faithful output, demonstrating the effectiveness of the proposed stepwise 2D personalization strategy in capturing details in the reference image.

We also compare our stepwise 2D personalization with another multiple object personalization method, i.e., Custom Diffusion. 
As shown in Fig.~\ref{fig_custom}, our stepwise 2D personalization achieves high-quality and faithful personalization results and ensures precise editing within the specified region.
In contrast, although Custom Diffusion assigns separate special text tokens to each \emph{concept}, the UNet is updated by all \emph{concepts}, resulting in less satisfactory personalization results.
In addition, Custom Diffusion lacks a localization mechanism to specify the interaction between the sunglasses and the toy, failing to place the new object on the toy's forehead.

\subsubsection*{Ablation study on different 3D representations}
We test different 3D representations in Fig.~\ref{fig_ablation_learning_2} and Tab.~\ref{tab:ablation_3D_repr} while keeping all the other settings the same.
Using GS obtains the best editing result while keeping the background unchanged.
For Instant-NGP~\cite{muller2022instant}, we observe undesired changes in the background since its content in different locations is not independent due to its adoption of a shared MLP and multi-resolution grid.
For NeuMesh~\cite{yang2022neumesh} (used in DreamEditor), we observe obvious artifacts if the editing requires (relatively) large shape change since it adopts mesh as its geometry proxy, which is less flexible than GS due to topology constraints.

\begin{figure*}
\includegraphics[width=0.95\textwidth]{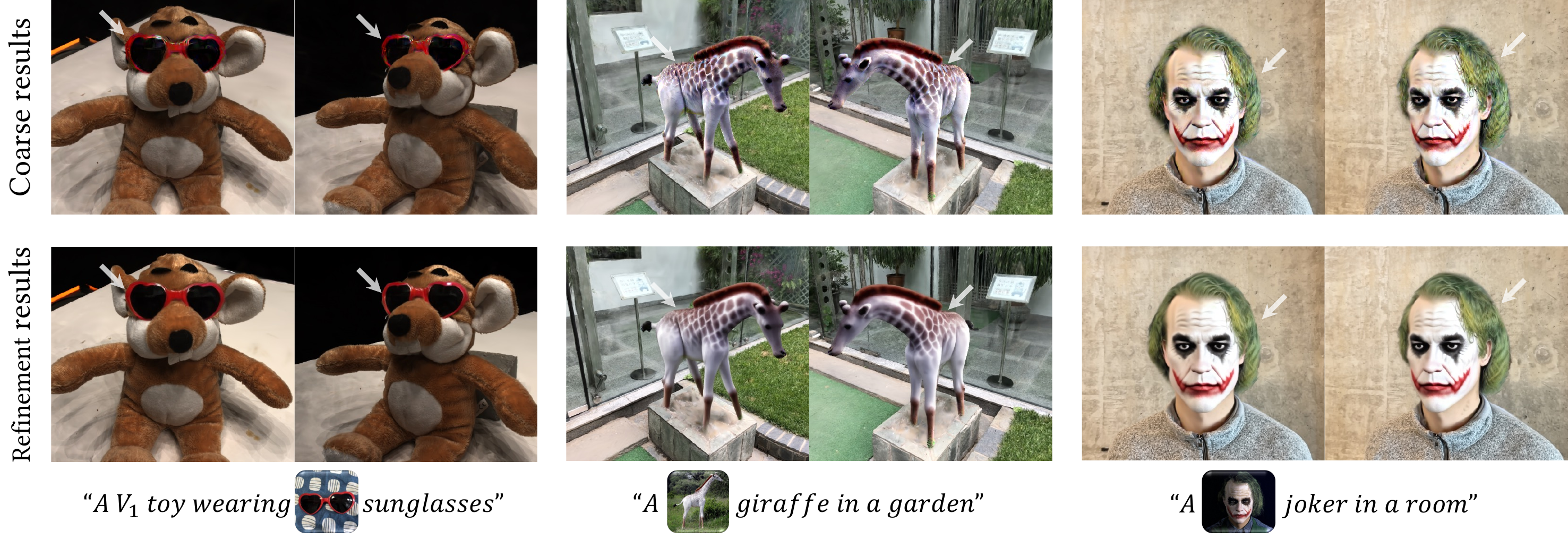}
\caption{
Comparison of the coarse editing results and the refinement results.
The region indicated by the arrow demonstrates the efficacy of the refinement step in enhancing the quality of the editing results.
}
\label{fig:corase_res}
\end{figure*}

\subsubsection*{Influence of different $\gamma$ in coarse editing}
As in Fig.\ref{fig_ablation_learning_3}, both the global and local SDS loss are necessary and our solution achieves the best result.
Specifically, only using global SDS loss $\mathcal{L}^{G}_{SDS}$ results in obvious artifacts in the editable region.
Only using local SDS loss $\mathcal{L}^{L}_{SDS}$ results in inaccurate placement of the object and unnatural color discrepancy between the background and the novel content since the context information is missing during optimization.

\subsubsection*{Effectiveness of the pixel-level refinement step}
As in Fig.\ref{fig:corase_res}, introducing the refinement stage effectively reduces artifacts and enhances the texture, resulting in substantially improved quality.

\begin{figure}
\includegraphics[width=\columnwidth]{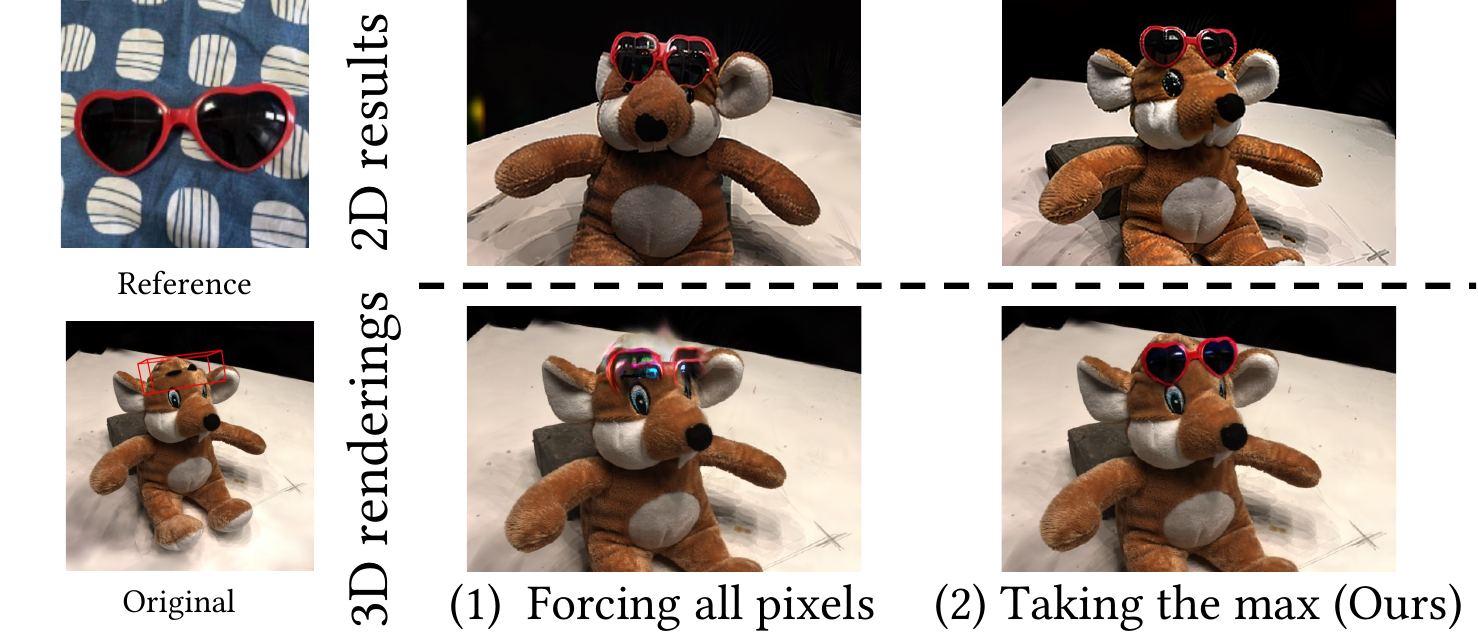}
\caption{
Ablation study on different formulas of $\mathcal{L}_{loc}$.
We compare the generated images of the personalized T2I model (top row) and the rendered images of the updated 3D scene (bottom row). 
Forcing all attention values in the region $\bar{\mathcal{S}}$ to 1 tends to create an object (even multiple objects) covering the entire bounding box regardless of whether its placement is appropriate (e.g. overfill), resulting in various artifacts.
}
\label{fig_ablation_loc}
\end{figure}

\subsubsection*{Ablation study on different formulas of $\mathcal{L}_{loc}$}

In Eq.~\ref{equ:loss_att}, we only encourage the max attention value of the pixels in the region $\bar{\mathcal{S}}$ to approach 1.
This is different from previous works~\cite{xiao2023fastcomposer,avrahami2023break}, which force all the attention values in the region $\bar{\mathcal{S}}$ to 1.
We compare these two settings in Fig.~\ref{fig_ablation_loc}.
Using the latter one tends to create an object (even multiple objects) covering the entire bounding box regardless of whether its placement is appropriate (e.g. overfill), resulting in various artifacts.

\section{Conclusion and Limitations}

In this paper, our proposed \sysname{} equips the emerging text-driven 3D editing with an additional image prompt as a complement to the textual description and produces high-quality editing results accurately aligned with the text and image prompts while keeping the background unchanged.
\sysname{} offers significantly enhanced controllability and enables versatile applications, including object insertion, object replacement, re-texturing, and stylization.

One limitation of \sysname{} is the coarse bounding box input.
Although convenient, it struggles in complex scenes where bounding boxes may include unwanted elements. 
It would be very beneficial to automatically obtain 3D instance segmentation of the scene.
Another limitation is related to geometry extraction since it is hard to extract a smooth and accurate mesh from GS-represented scenes.

\begin{acks}
This work was supported in part by the National Natural Science Foundation of China (NO.~62325605, NO.~62322608), in part by the CAAI-MindSpore Open Fund, developed on OpenI Community,in part by the Open Project Program of State Key Laboratory of Virtual Reality Technology and Systems, Beihang University (No.VRLAB2023A01).
\end{acks}

\bibliographystyle{ACM-Reference-Format}
\bibliography{TIEditor}

\end{document}